\documentclass[]{interact}

\usepackage{epstopdf}
\usepackage[caption=false]{subfig}

\usepackage[numbers,sort&compress]{natbib}
\usepackage{fancyhdr}
\usepackage{amsmath}
\usepackage{amsfonts}
\usepackage{threeparttable}
\usepackage[mathscr]{eucal}
\usepackage{color}
\usepackage{array}
\usepackage{fmtcount}
\usepackage{geometry}
\usepackage{algorithm,algorithmic}
\usepackage{multicol}
\usepackage{listings}
\usepackage{caption}
\bibpunct[, ]{[}{]}{,}{n}{,}{,}
\makeatletter
\def\NAT@def@citea{\def\@citea{\NAT@separator}}
\makeatother

\theoremstyle{plain}
\newtheorem{theorem}{Theorem}[section]

\theoremstyle{definition}
\newtheorem{definition}[theorem]{Definition}

\theoremstyle{remark}
\newtheorem{remark}{Remark}

\begin{document}

\articletype{ARTICLE TEMPLATE}

\title{Weak Equitability of Dependence Measure}

\author{ $^{1}$\footnote{Correspondence author: } \quad $^{2}$ \quad $^{3}$ }

\author{
\name{Hangjin Jiang\thanks{CONTACT H. Jiang. Email: jianghangjin10@mails.ucas.ac.cn}, Kan Liu and Yiming Ding}
\affil{Wuhan Institute of Physics and Mathematics, Chinese Academy of Sciences, Wuhan, China}
}

\maketitle

\begin{abstract}
Measuring dependence between two random variables is very important, and critical in many applied areas such as variable selection, brain network analysis. However, we do not know what kind of functional relationship is between two covariates, which requires the dependence measure to be equitable. That is, it gives similar scores to equally noisy relationship of different types. In fact, the dependence score is a continuous random variable taking values in $[0,1]$, thus it is theoretically impossible to give similar scores. In this paper, we introduce a new definition of equitability of a dependence measure, i.e, power-equitable (weak-equitable) and show by simulation that HHG and Copula Dependence Coefficient (CDC) are weak-equitable.
\end{abstract}


\section{Introduction}
Measuring dependence between two random variables plays a fundamental role in various kind of data analysis, such as fMRI data, genetic data. How should one quantify such a dependence without bias for relationship of a specific form?  This gives rise to the concept ``Equitability" of a dependence measure \cite{mic, equal-mic}. By scoring relationships according to an equitable measure one hopes to find important patterns of any type. The first description of ``Equitability" given in \cite{mic} is ``A measure of dependence is said to be equitable if it gives similar scores to equally noisy relationship of different types". In \cite{equality}, authors pointed out that there is no dependence satisfying the 
definition of equitability given in \cite{mic}, the $R^2$-Equitability. Furthermore, they give a definition of ``self-equitability", and show by theoretical result and numerical simulation that mutual information satisfies the ``self-equitability".

Equitability of dependence measure means it is equitable to all kinds of functional relationship,  as the description goes, give similar scores to equally noisy functional relationships. 
Taking correlation for example, it is un-equitable, since it gives very small scores to 
nonlinear relationships that can not be well approximated by linear function.
There are two key terms in this sentence  ``equally noisy relationship" and ``similar score".  
The noise level is given by $1-R^2\{f(X),Y\}$ for the model $Y=f(X)+\epsilon$ in \cite{equal-mic, equality}, where $Y$ is a continuous response, $X$ is a continuous covariate, and $\epsilon$ is the noise term independent of $X$. In this paper, we introduce the model signal-to-noise ratio (MSNR) to control the noise level in our simulation, and give the definition of equitability in a more natural way.

The self-equitablility defined and discussed in \cite{equality} focuses on the equitability of dependence measure when it is regarded as a measure of noise in the data set. If a dependence measure, $D$, is self-equitable, it does not depend on what the specific functional relationship between $X$ and $Y$, i.e, $D(X,Y)=D(f(X),Y)$, where $f$ is a Boreal measurable function such that $X\leftrightarrow f(X) \leftrightarrow Y$ forms a Markov chain.
However, the noise in the data set is difficult to measure, or no one can tell through existing methods how much noise is in the data set. In other words, self-equitable is necessary for a dependence measure used as a noise measure.  However, in practice, people usually used the dependence measure as a test statistic. That is, we say $X$ and $Y$ are significantly associated with each other when the value $D[X;Y]$ obtained by dependence measure $D$ is larger than a given threshold, which motivated us to the definition of power-equitable. Since if a dependence measure is equitable, it is power-equitable. 
Thus, we also call it weak-equitable.

In this paper, we try to make clear the statement debates in \cite{debate} that MIC is more equitable that MI. In addition, we introduce a new definition, weak-equitable (power-equitable), which is meaningful when a dependence measure is used to test independence.  In Section 2, we recall firstly the definitions given in \cite{equality}, and then introduce the definition of equitable, and weak-equitable.  Simulation results are given in section 3.
\section{Definitions of Equitability}

A measure of dependence is said to be equitable if it gives similar scores to equally noisy relationships of different types \cite{mic, equal-mic}. In other words, a measure of how much noise is in an $x$-$y$ scatter plot should not depend on what the specific functional relationship between $x$ and $y$ would be in the absence of noise \cite{equality}. Justin B. Kinney et al. gave the definition of $R^2$-equitable and self-equitable \cite{equality}. They pointed out that the dependence measure satisfying $R^2$-equitable does not exist, and dependence measure satisfying self-equitability exists, mutual information is one of them.

In the following, we recall the definition of self-equitable and $R^2$-equitable.
\begin{definition}
	A dependence measure $D[X;Y]$ is $R^2$-equitable if and only if, when evaluated on a joint probability distribution $P(X,Y)$, that corresponds to a noisy functional relationship between two real random variables $X$ and $Y$, the following relation holds:
	\begin{equation}
	D[X;Y]=g(R^2[f(X);Y])
	\end{equation}
	Here, $g$ is a function that does not depend on $P(X,Y)$ and $f$ is the function defining the noisy functional relationship, i.e., $Y= f(X)+\eta$,  for some random variable $\eta$. The noise term $\eta$ may depend on $f(X)$  as long as $\eta$ has no additional dependence on $X$, i.e, $x \rightarrow f(x)\rightarrow \eta$ forms a Markov chain.
	
\end{definition}
\begin{definition}\label{equitable}
	A dependence measure $D[X; Y]$ is self-equitable if and only if
	\begin{equation}
	D[X; Y] = D[f(X); Y]
	\end{equation}
	whenever $f$ is a deterministic function and $X \rightarrow f(X)\rightarrow Y$ forms a Markov chain.
\end{definition}

In the definition of $R^2$-equitable, the term ``gives similar scores to equally noisy relationships of different types" is described by the dependence measure as a function, independent with $X$ and $Y$, of noise, so the meaning of equitability is conveyed implicitly.  Differently, our definition of equitable conveys the meaning, ``equally noisy relationship",  directly through a mathematical definition of noisy-equal model based on the model signal-to-noise ratio (MSNR).

\begin{definition} (MSNR) Signal-to-Noise Ration for a model, $Y=f(X)+\varepsilon$, is given by
	\begin{equation}
	\text{MSNR}_{\varepsilon}(f)=\frac{\text{var}(Y)}{\text{var}(\varepsilon)}
	\end{equation}
	where $\text{var}(X)$ is the variance of $X$, $\varepsilon$ is the noise term assumed to be normal distributed $N(0,\sigma^2)$.
\end{definition}
 In the following, two models,  $Y_1=f_1(X)+\varepsilon_1$ and $Y_2=f_2(X)+\varepsilon_2$ with the same MSNR are called \textbf{noisy-equal models}.
\begin{remark}
	
\ \

\begin{itemize}
	\item In \cite{mic, equality}, the noise level is measured by 1-$R^2(f(X),f(X)+\varepsilon)=1-\rho^2(f(X),f(X)+\varepsilon)$. We have
	\begin{equation}
	R^2(f(X),f(X)+\varepsilon)=\frac{1}{\sqrt{1+\frac{1}{MSNR_{\varepsilon}(f)}}}
	\end{equation}
	\item How to get noisy-equal models? Given two models,   $Y_1=f_1(X)+\varepsilon_1$ and $Y_2=f_2(X)+\varepsilon_2$, we set $\varepsilon_1=\varepsilon$, $\varepsilon_2=\frac{\varepsilon}{a}$, where $a^2=\frac{var(f_1(X))}{var(f_2(X))}$, $\varepsilon$  is independent with $X$, then
	\begin{equation}\label{cmsnr}
	\begin{split}
	{\rm MSNR}_{\varepsilon_1}(f_1)  &=  \frac{{\rm var}(Y_1)}{{\rm var}(\varepsilon)} = \frac{{\rm var}(f_1(X))+{\rm var}(\varepsilon)}{{\rm var}(\varepsilon)}\\
	&=  \frac{{\rm var}(f_1(X))/a^2+{\rm var}(\varepsilon/a)}{{\rm var}(\varepsilon/a)} \\
	&= \frac{{\rm var}(Y_2)}{{\rm var}(\varepsilon_2)} = {\rm MSNR}_{\varepsilon_2}(f_2)
	\end{split}
	\end{equation}
	Hence, we get noisy-equal models $Y_1=f_1(X)+\varepsilon$ and $Y_2=f_2(X)+\frac{\varepsilon}{a}$, where $a^2=\frac{{\rm var}(f_1(X))}{{\rm var}(f_2(X))}$.
\end{itemize}
\end{remark}

\begin{definition}\label{equitable}(Equitable)
	A dependence measure $D[X; Y]$ is Equitable if and only if
	\begin{equation}\label{equal-def}
	D[X; Y_1] = D[X; Y_2]
	\end{equation}
	for any noisy-equal models  $Y_1=f_1(X)+\varepsilon_1$ and $Y_2=f_2(X)+\varepsilon_2$.
\end{definition}

This definition theoretically equals to that of $R^2$-equitable presented in \cite{equality}, 
however, it is more natural and heuristic. Furthermore, it helps us to look insight into equitability of dependence measures as shown in the following  section.

The self-equitable focused on the equitability of dependence measure when it is regarded as a measure of noise in the data set.  If a dependence measure is self-equitable, then it does not depend on what the specific functional relationship between $X$ and $Y$ would be in the absence of noise. However, in practice, people usually used the dependence measure as a test statistic, which motivated us to the definition of power-equitable, that is, weak-equitable. 

\begin{definition}\label{equitable}(Power-equitable)
	A dependence measure $D[X; Y]$ is Power-equitable if and only if
	\begin{equation}
	{\rm Power}[D,f_1] = {\rm Power}[D,f_2]
	\end{equation}
	for any noisy-equal models  $Y_1=f_1(X)+\varepsilon_1$ and $Y_2=f_2(X)+\varepsilon_2$, where Power$[D,f]$ is the power of dependence measure $D$ in detecting functional relationship $f$ underlying noisy data set.
\end{definition}

So, if a dependence measure is equitable, then it is power-equitable. The converse is not right. We refer the power-equitable as weak-equitable.

Weak-equitability can be explained as follows. If a dependence measure $D$ is weak-equitable,  all kinds of functional relationship underlying equally noisy data sets will be detected with the same possibility by $D$, which is quite meaningful and interesting.  This allowed, similar to a equitable case, us to use $D$ as a test to determine whether or not $X$ and $Y$ is significantly associated with each other.  As mentioned before, self-equitable is meaningful, when $D$ is used as a noise measure.

\section{Simulation Results}
In this section, we discuss the self-equitability, equitability, and power-equitability of some popular dependence measures: MIC\cite{mic}, CDC\cite{review}, RDC\cite{RDC}, HHG\cite{hhg}, Pearson Correlation coefficient (pcor), Spearman's Rank Correlation (scor), Kentall's $\tau$ (kcor), curve correlation\cite{curvecor},  HSIC\cite{hsdm}, normalized mutual information(MI)\cite{mic}.  A detailed discussion of these measures can found in \cite{review}. The equitability of dcor\cite{dcor} is discussed in \cite{equal-mic} and that of MIC is discussed in \cite{equal-mic}.

\subsection{Simulation settings}
In our simulation, (a) we sample $X$ from uniform distribution with length $n=1000$; (b) $\varepsilon$ is sampled from standard normal distribution with the same sample size as $X$; and (c) $Y$ is generated according to one of 21 different kinds of relationships given in Appendix. To get a sample of $D(X,Y)$ with sample size $N$ for a given functional relationship, repeat (a-c) for $N$ times. We set $N=100$ for equitability simulation, and $N=300$ for power/weak-equitability simulation.

\subsection{Equitability}

In this part, we analyze the equitability of MIC, CDC, RDC, HHG, pcor, scor, kcor, dcor, HSIC and MI.  Although, there is some theoretical results given in \cite{equality} showing that dependence measure satisfying the definition of equitable does not exist, we give a detailed  simulation results for further explanation.

\begin{figure}[!htbp]
	\begin{center}
		\includegraphics[width=5.5in, height=5.3in]{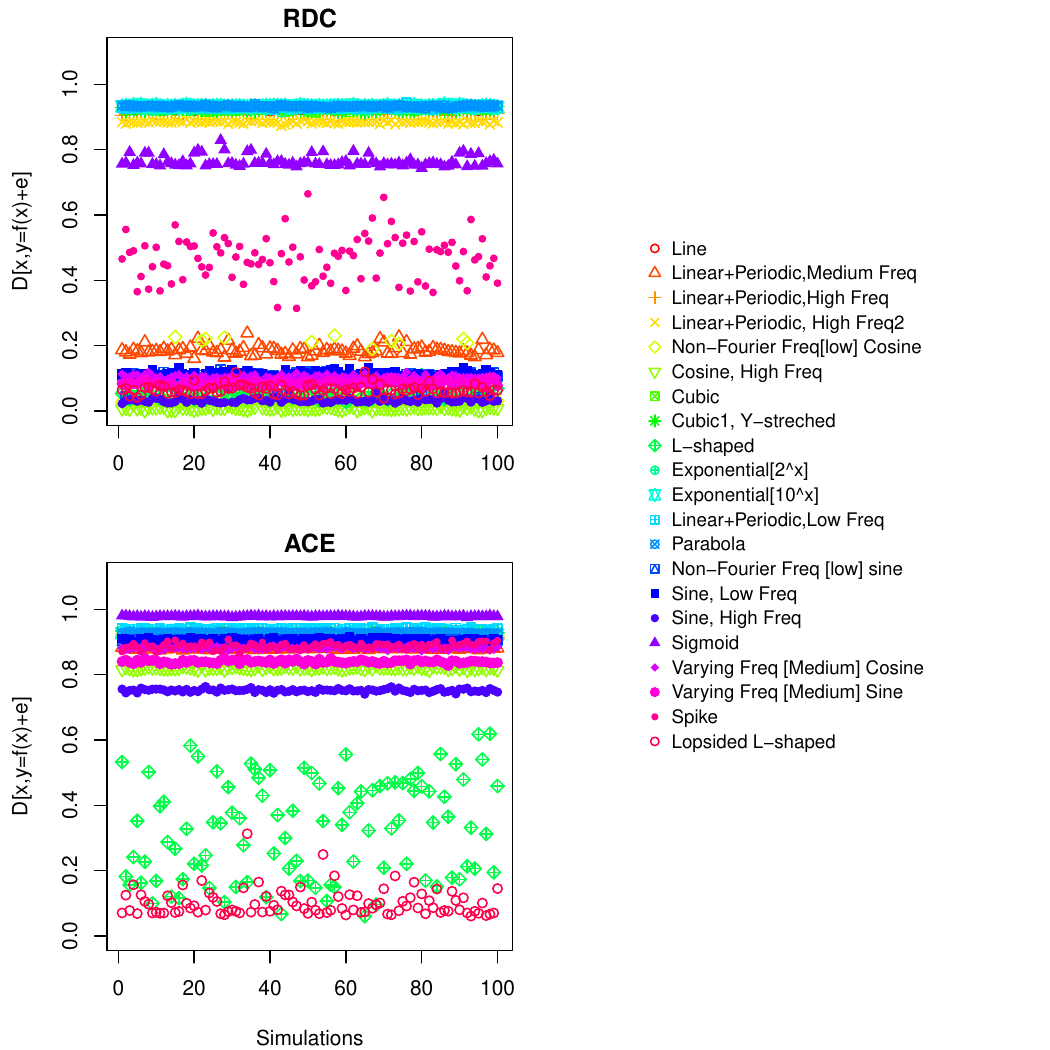}
	\end{center}
	\caption{(color online) Equitability of ACE and RDC with mean(MSNR)=11.529, sd(MSNR)=0.233. 
		The simulation is based on 21 types of relationship (Definitions are given in Appendix). MSNR is controlled using (\ref{cmsnr}). According to the definition of equitability, they are both not equitable. In addition, RDC gives different scores to functional type `Spike' in 100 simulations, and ACE gives different scores to `L-shaped'.}
	\label{equal1}
\end{figure}

\begin{figure}[!htbp]
	\begin{center}
		\includegraphics[width=5.5in, height=5.5in]{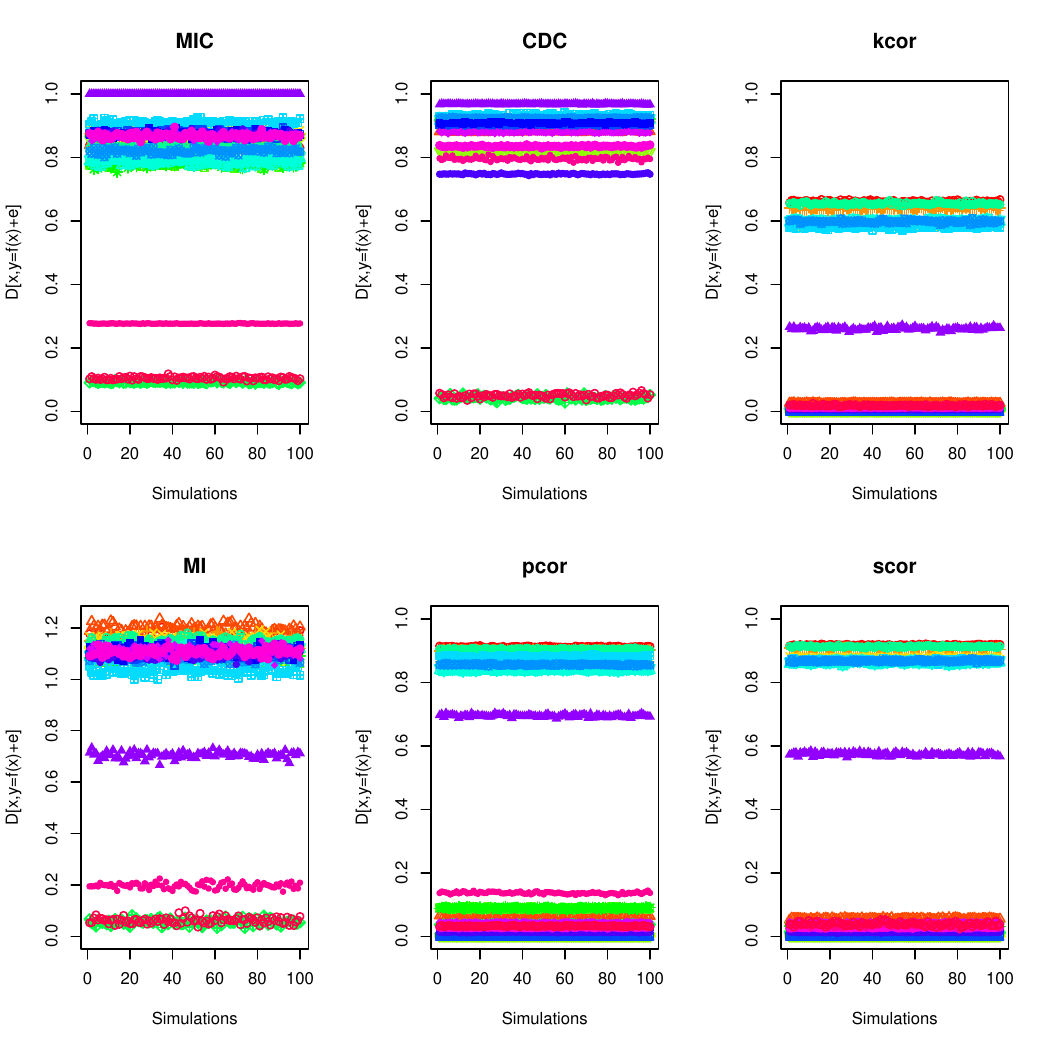}
	\end{center}
	\caption{(color online) Equitability of MIC, MI, CDC, pcor, kcor, scor with mean(MSNR)=11.529, sd(MSNR)=0.233 (Legends are the same as that in Figure 1).
		The simulation is based on 21 types of functional relationships (Definitions are given in Appendix). MSNR is controlled using (\ref{cmsnr}).  According to the definition of equitability, these methods are not equitable. MI is equitable in a small set of functional types except four types: Sigmoid, Lopsided L-shaped, L-shaped and Spike. The performance of MIC is similar to MI, its values on these functional types except four of them are located in a small range. Especially, MIC values locate in $[0.78,0.94]$, and MI values in $[1,1.21]$.}
	\label{equal6}
\end{figure}

Figure \ref{equal1} shows the simulation results of ACE, and RDC.
According to the definition of equitability (Definition \ref{equitable}), we can see that both RDC and ACE are not equitable. Figure \ref{equal6} shows the simulation results of MIC, MI, CDC, poor, kcor and scor. Similarly, all of them are not equitable. Interestingly, the performance of MIC and MI are quite similar.

\subsection{Weak-Equitable}

\begin{figure}[!htbp]
	\begin{center}
		\includegraphics[width=5.5in, height=5.3in]{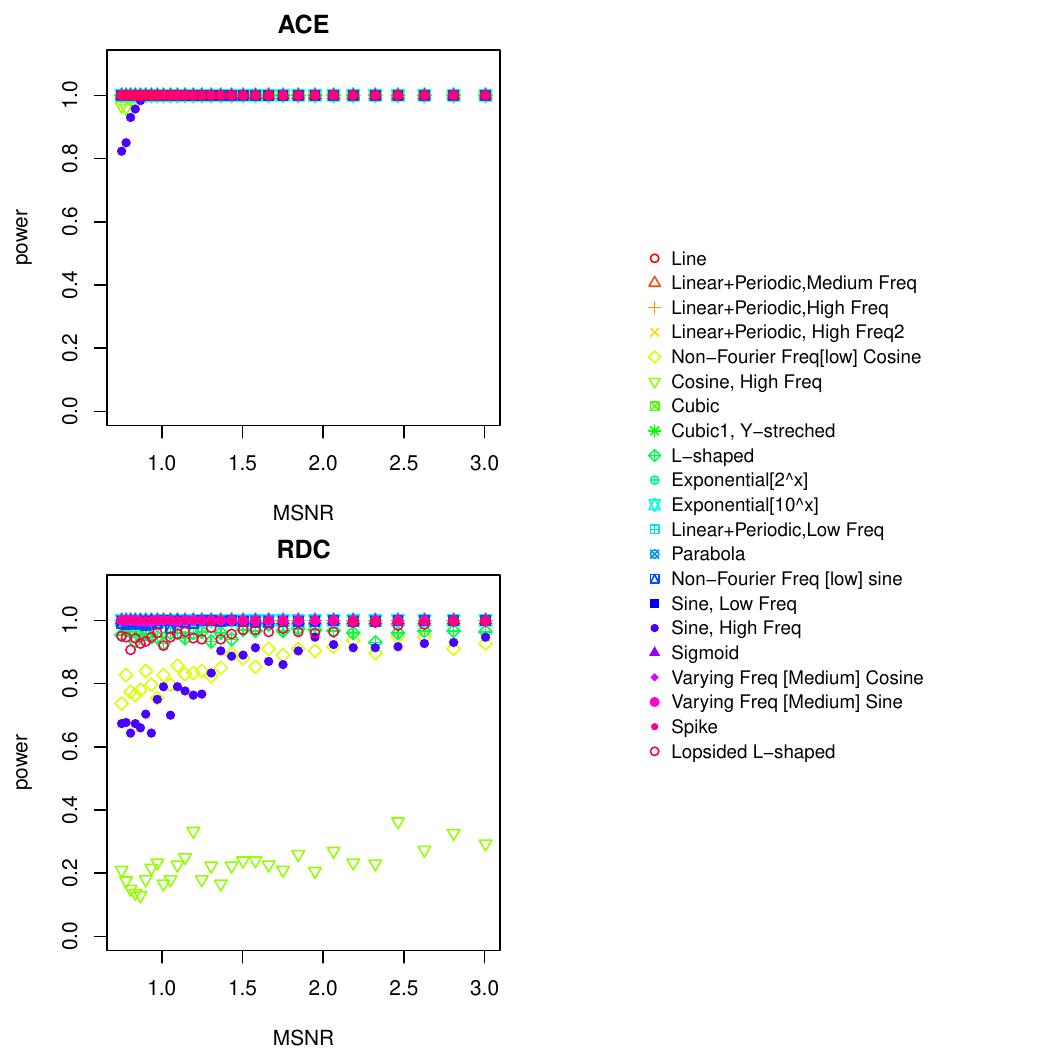}
	\end{center}
	\caption{(color online) Weak-Equitable of ACE and RDC with mean MSNR ranges from 0.75 to 3
		(Legends are the same as that in Figure 1).
		The simulation is based on 21 types of relationship (Definitions are given in Appendix). MSNR is controlled using (\ref{cmsnr}).  We can see from the results that ACE is much more equitable than RDC in this range of MSNR. 
	}
	\label{wequal1}
\end{figure}

\begin{figure}[!htbp]
	\begin{center}
		\includegraphics[width=5.5in, height=5.3in]{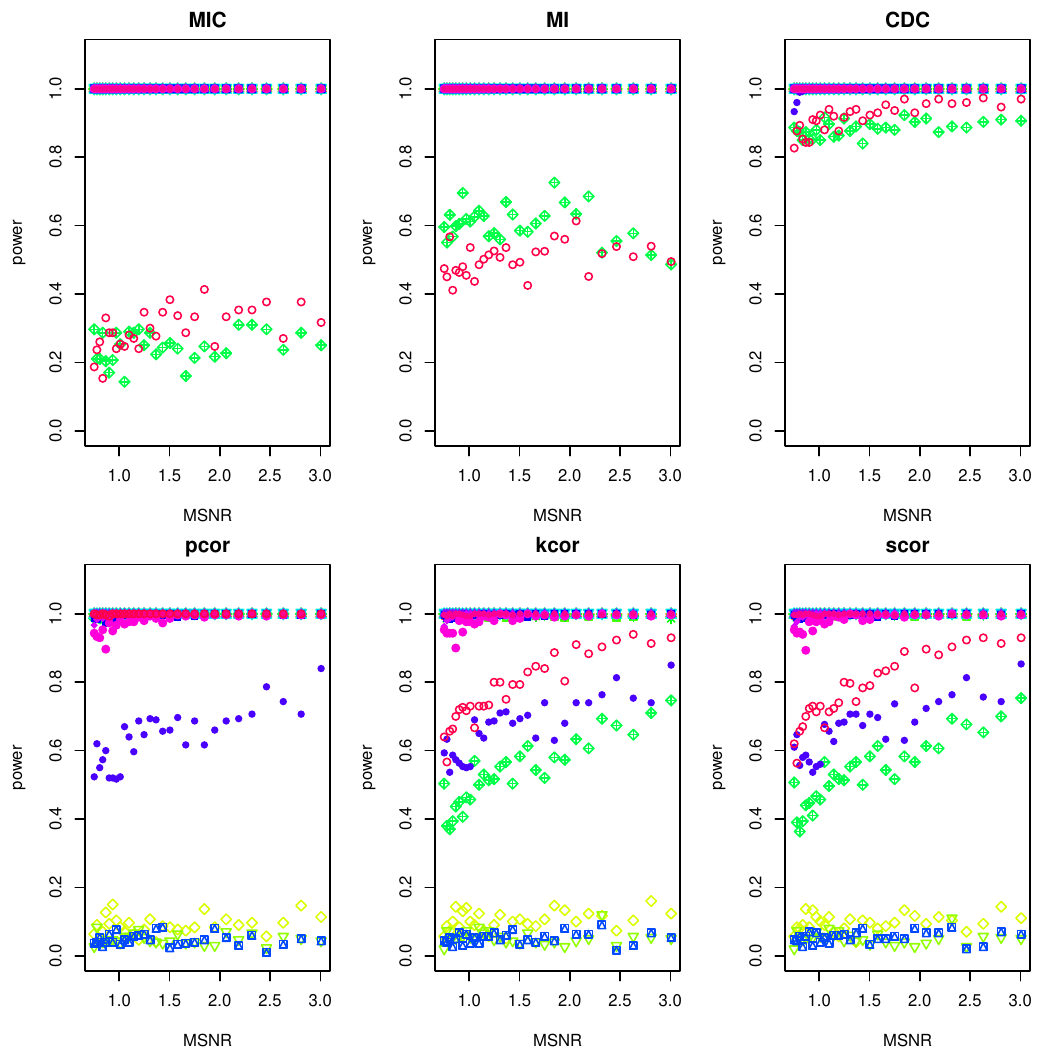}
	\end{center}
	\caption{(color online) Weak-Equitable of  MIC, MI, CDC, pcor, kcor and scor with mean MSNR range from 0.75 to 3 (Legends are the same as that in Figure 1).
		The simulation is based on 21 types of relationship (Definitions are given in Appendix). MSNR is controlled using (\ref{cmsnr}). We can see from the results that CDC is much more equitable than others in this range of MSNR. For MI and MIC, their power-equitable property is very similar. It seems that MI is more powerful than MIC, since MI has a higher power on L-shaped and Lopsided L-shaped than MIC.
	}
	\label{wequal2}
\end{figure}

\begin{figure}[!htbp]
	\begin{center}
		\includegraphics[width=5.5in, height=2.8in]{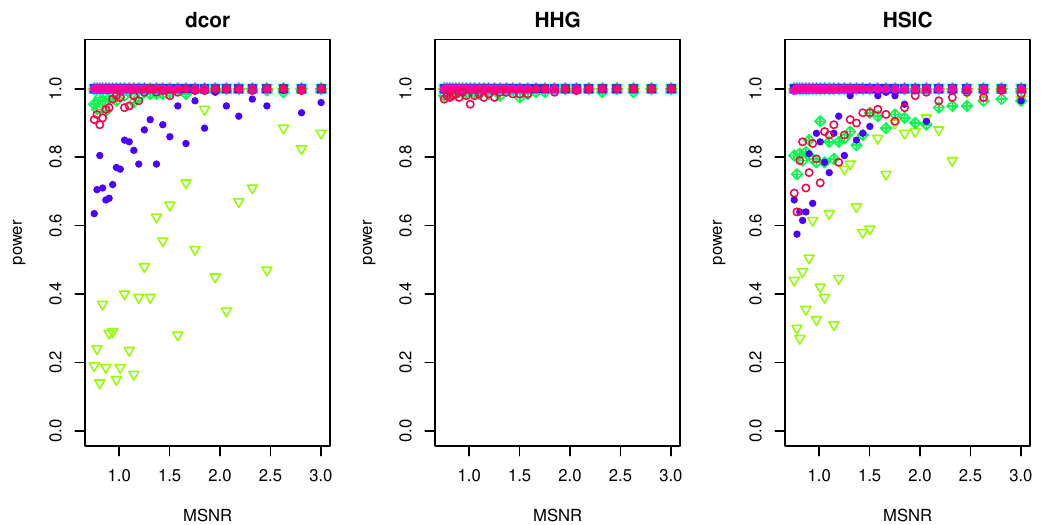}
	\end{center}
	\caption{(color online) Weak-Equitable of dcor, HHG, and HSIC with mean MSNR range from 0.7509836 to 3.0057
		(Legends are the same as that in Figure 1).
		The simulation is based on 21 types of relationship (Definitions are given in appendix), MSNR is controlled using (\ref{cmsnr}),and simulation details are given in section 3.1.  We can see from the results that HHG is much more equitable than others in that range of MSNR, here we emphasize on the range of MSNR because if there is a very small MSNR for the data, then all of the methods will have a very small power.
	}
	\label{wequal3}
\end{figure}

In this part, we discussed the weak-equitable of some popular dependence measures.  In our simulation, we set MSNR ranging from 0.75 to 3. The results given in Figure \ref{wequal1}, Figure \ref{wequal2},  and Figure \ref{wequal3} show that HHG is almost power-equitable, and ACE is secondary to HHG. However ACE is sensitive to outliers, we recommend to use CDC in large data sets for association detecting.

\section{Conclusion}

In this paper, we discussed  the equitability and weak-equitability (power-equitability) of some common dependence measures, such as MIC, MI, dcor, CDC, ACE, RDC, HHG, Pearson correlation coefficient (pcor), Spearman rank correlation (scor), kendall's $\tau$, and HSIC.  The equitability requires a dependence measure giving similar scores to equally noisy data sets no matter what kind of functional relationship is underlying them, this requirement is so strong that it can not be satisfied by any dependence measure. Therefore, we introduced power-equitability, which only requires that all kinds of functional relationship, linear or non-linear, can be detected with the same possibility by the dependence measure. It is also called as weak-equitability because if $D$ is equitable, then it is power-equitable.

Self-equitable is the basic requirement for a dependence measure to be used as a noise measure. However, a self-equitable measure, such as MI, may not have a satisfying power in finding relationships as shown in our simulation results (see Figure \ref{wequal2}).

Based on our simulation, we find that MI is more equitable than MIC, HHG is power-equitable, and ACE/CDC is secondary to HHG.

\section{Appendix}

\subsection{Definition of functions}
\label{def4func}
\begin{enumerate}
	\item Line: $y=x$
	\item Linear+Periodic, Low Freq: $y=0.2 \sin(4(2x-1))+\frac{11}{10}(2x-1)$
	\item Linear+Periodic, Medium Freq: $y=\sin(10\pi x) +x$
	\item Linear+Periodic, High Freq:  $y=0.1 \sin(10.6(2x-1))+\frac{11}{10}(2x-1)$
	\item Linear+Periodic, High Freq:  $y=0.2 \sin(10.6(2x-1))+\frac{11}{10}(2x-1)$
	\item Non-Fourier Freq [Low] Cosine: $y=\cos(7\pi x)$
	\item Cosine, High Freq: $y=\cos(14 \pi x)$
	\item Cubic: $y=4x^3+x^2-4x$
	\item Cubi, Y-stretched: $y=41(4x^3+x^2-4x)$
	\item L-shaped: $y=x/99I(x \leq \frac{99}{100}) + I(x>\frac{99}{100})$
	\item Exponential [$2^x$] : $y=2^x$
	\item Exponential [$10^x$] : $y=10^x$
	\item Parabola: $y=4x^2$
	\item Non-Fourier Freq [Low] Sine: $ y=\sin(9\pi x)$
	\item Sine, Low Freq:  $y=\sin(8\pi x)$
	\item Sine, High Freq: $ y=\sin(16 \pi x)$
	\item Sigmoid: $y=[50(x-0.5)+0.5]I(\frac{1}{20} \leq x \leq \frac{51}{100}) + I(x>\frac{51}{100})$
	\item Varying Freq [Medium] Cosine: $y=\sin(5\pi x(1+x))$
	\item Varying Freq [Medium] Sine: $ y=\sin(6\pi x(1+x))$
	\item Spike: $y=20I(x<\frac{1}{20}) +[-18x+\frac{19}{10}]I(\frac{1}{20} \leq x < \frac{1}{10}) + [-\frac{x}{9}+\frac{1}{9}]I(x \geq \frac{1}{10})$
	\item Lopsided L-shaped: $y=200xI(x<\frac{1}{200}) + [-198x+\frac{199}{100}]I(\frac{1}{200} \leq x < \frac{1}{100}) + [-\frac{x}{99}+\frac{1}{99}]I(x \geq \frac{1}{100})$
	
\end{enumerate}
where $I(\cdot)$ is the indicator function.

%

\end{document}